\documentclass[letterpaper, 10 pt, conference]{ieeeconf} 
\IEEEoverridecommandlockouts                             
\usepackage{algorithm}
\usepackage{algpseudocode}
\usepackage{amsfonts}
\usepackage{amsmath}
\usepackage{amssymb}
\usepackage{booktabs}
\usepackage{comment}
\usepackage[T1]{fontenc}
\usepackage[utf8]{inputenc}
\usepackage{mathtools}
\usepackage{nicefrac}
\usepackage{setspace}
\usepackage[separate-uncertainty=true,detect-mode=true, group-separator = {,}, group-minimum-digits=3]{siunitx}
\usepackage{url}

\usepackage{color}
\usepackage{graphics}
\usepackage{pgfplots}
\pgfplotsset{compat=newest}
\pgfplotsset{every axis legend/.append style={legend cell align=left}}
\usepackage[mode=buildnew]{standalone}
\usepackage{subcaption}
\usepackage{tikz}

\usetikzlibrary{matrix}
\usetikzlibrary{calc, positioning}
\usetikzlibrary{arrows.meta}

\newcommand\dd{0.7cm} 
\tikzset{
  >=latex,
  mybox/.style={
    rectangle,
    fill=white,
    thin,
    draw, outer sep=0,
    minimum width = \dd,
    minimum height = \dd,
    inner sep=0,
    node distance = \dd,
  },
}

\definecolor{lr}{RGB}{255,185,177}
\definecolor{lg}{RGB}{169,219,209}
\definecolor{lb}{RGB}{213,237,249}
\definecolor{tan}{RGB}{255,225,185}
\definecolor{obcolor}{RGB}{220,223,240}
\usetikzlibrary{shapes,arrows}

\definecolor{emb_color}{RGB}{255,185,177}
\definecolor{multi_head_attention_color}{RGB}{255,225,185}
\definecolor{add_norm_color}{RGB}{242,243,193}
\definecolor{ff_color}{RGB}{213,237,249}
\definecolor{softmax_color}{RGB}{169,219,209}
\definecolor{linear_color}{RGB}{220,223,240}
\definecolor{gray_bbox_color}{RGB}{243,243,244}

\pgfplotsset{
    layers/my layer set/.define layer set={
    background,
    pre main,
    main,
    trajectory,
    reddots,
    ellipses,
    foreground,
    }{},
    set layers=my layer set,
}

\definecolor{viridis1}{RGB}{227, 228, 18}
\definecolor{viridis2}{RGB}{31, 161, 135}
\definecolor{viridis3}{RGB}{70, 50, 127}
\definecolor{cardinal}{RGB}{140, 21, 21}
\definecolor{fire_opal}{RGB}{223, 96, 71}
\definecolor{dark_red}{RGB}{220, 0, 0}

\definecolor{vandeusen}{RGB}{73,92,111}
\definecolor{cordovan}{RGB}{152,68,71}
\definecolor{pastelBlue}{RGB}{0,114,178}
\definecolor{pastelRed}{RGB}{245,97,92}
\definecolor{pastelGreen}{RGB}{0,158,115}
\definecolor{pastelPurple}{RGB}{135,112,254}
\definecolor{pastelOrange}{RGB}{230,159,0}
\definecolor{pastelSkyBlue}{RGB}{86,180,233}
\definecolor{darkbay}{HTML}{417865}
\usepackage[colorlinks = true,
            linkcolor = darkbay,
            urlcolor  = darkbay,
            citecolor = darkbay,
            anchorcolor = red]{hyperref}

\usepackage[style=ieee,backend=biber,url=false,doi=false,natbib=true,mincitenames=1,dashed=false,isbn=false]{biblatex}
\usepackage[capitalize]{cleveref}
\AtEveryBibitem{\clearfield{issn}}
\AtEveryCitekey{\clearfield{issn}}

\DeclarePairedDelimiterX{\infdivx}[2]{(}{)}{%
  #1\;\delimsize\|\;#2%
}

\title{\LARGE \bf
Deep Normalizing Flows for State Estimation
}

\author{Harrison Delecki\textsuperscript{*}, Liam A. Kruse\textsuperscript{*}, Marc R. Schlichting\textsuperscript{*}, and Mykel J. Kochenderfer
\thanks{\textsuperscript{*} These authors contributed equally}%
\thanks{H. Delecki, L.A. Kruse, M.R. Schlichting, and M.J. Kochenderfer are with the Stanford Intelligent Systems Laboratory in the Department of Aeronautics and Astronautics at Stanford University, Stanford, CA 94305, USA (email: \{hdelecki, lkruse, mschl, mykel\}@stanford.edu).
}
}

\begin{document}
\maketitle
\thispagestyle{empty}
\pagestyle{empty}

\begin{abstract}
Safe and reliable state estimation techniques are a critical component of next-generation robotic systems. Agents in such systems must be able to reason about the intentions and trajectories of other agents for safe and efficient motion planning. However, classical state estimation techniques such as Gaussian filters often lack the expressive power to represent complex underlying distributions, especially if the system dynamics are highly nonlinear or if the interaction outcomes are multi-modal. In this work, we use normalizing flows to learn an expressive representation of the belief over an agent's true state. Furthermore, we improve upon existing architectures for normalizing flows by using more expressive deep neural network architectures to parameterize the flow. We evaluate our method on two robotic state estimation tasks and show that our approach outperforms both classical and modern deep learning-based state estimation baselines.
\end{abstract}

\section{Introduction}
State estimation is a crucial task in robotics applications such as autonomous driving and spacecraft attitude determination \cite{barfoot2017state}.
Examples of state variables include position, velocity, and the orientation of the robot. System measurements and robot dynamics might be stochastic in nature, leading to uncertainty about the underlying state. 
The core state estimation task is to infer the distribution over state $\mathbf{x}_t$ at time $t$ given observations $\mathbf{o}_{1:t}$. 
Unfortunately, classical state estimation techniques often lack the expressive power to represent complex distributions, especially if the system dynamics are nonlinear or if outcomes are multi-modal.%

Normalizing flows are a deep generative method for constructing probabilistic models of complex distributions using parametric variable transformations~\cite{papamakarios2021normalizinginference}. Transforms can be parameterized by simple deep neural networks, invertible neural networks, or recurrent networks. Normalizing flows have been used to represent high-dimensional and multi-modal distributions in image generation and time-series modeling~\cite{kobyzevintroductionmethods}. In this project we seek to improve upon existing architectures for normalizing flows by using more expressive deep neural network architectures. Furthermore, we apply our deep normalizing flow framework to robotics environments, specifically focusing on driving tasks---an area that has received little attention in the normalizing flow literature thus far. Furthermore, we compare different deep recurrent architectures for modeling sequential observations, including a novel transformer-based architecture to learn embeddings for normalizing flows.

\section{Related Work}
Classical filtering techniques such as the Kalman filter \cite{kalman1960new} are well-suited for applications in which unimodal Gaussian distributions represent the state distributions. To handle non-linear system dynamics, extensions such as the extended Kalman filter (EKF) and the unscented Kalman filter (UKF) \cite{julier1997new} have been introduced. However, many complex systems---especially highly stochastic multi-agent systems---induce multi-modal distributions over outcomes for which classical Gaussian filtering techniques fail. 

Given the potential multi-modality of the distribution $p(\mathbf{x})$, we must consider more advanced models for density estimation such as Gaussian mixture models (GMM) or mixture density networks (MDN) \cite{bishop1994mixture}. Mixture density networks (MDN) use deep neural networks to parameterize a conditional GMM that is optimized with a maximum likelihood objective. MDNs have been applied in domains such as human pose estimation \cite{li2019generating}, trajectory prediction \cite{Makansi_2019_CVPR}, and speech synthesis \cite{zen2014speech}. While they perform well in lower-dimensional spaces, they struggle with high-dimensional and non-Gaussian target distributions.

In recent years, deep generative models have been applied to state estimation. However, not all deep generative models provide explicit access to the probability density function. For example,  generative adversarial networks (GANs) \cite{goodfellow2020generative} can only sample from the latent distribution and do not explicitly compute its density. Variational autoencoders (VAEs) \cite{kingma2013auto} can provide access to the latent distribution while separately training an encoder and decoder. The training goal for VAEs has two components: learning a transform $Z=f(X)$ such that $Z\sim\mathcal{N}(\mathbf{0},\mathbf{I})$ and learning the approximate inverse of $f$. In the context of state estimation tasks, VAEs have been used for estimating and sampling the distribution of the quantum states of a system \cite{rocchetto2018learning} or for representing distributions over electroencephalogram measurements \cite{bethge2022eeg2vec}. However, VAEs often suffer from poor reconstruction errors because of the two-objective training goal.

The inherent issue with an encoder/decoder structure---namely, that the decoder is merely an approximate inverse of the encoder---is solved by the normalizing flow architecture, which learns a single invertible mapping. The interpretable yet highly expressive nature of normalizing flows has garnered interest in the robotics community. \citet{deng2020modeling} use normalizing flows to decode a base continuous stochastic process into a complex observable process. \citet{ma2020normalizing} represent multi-modal policies with conditional normalizing flows and demonstrate their effectiveness on a series of multi-agent games. Normalizing flows can also improve importance sampling. \citet{zhang2022accelerated} learn the distribution of risk events for autonomous vehicles and fit an importance sampling distribution with a masked autoregressive flow. \citet{zanfir2020weakly} conduct human pose and shape estimation using kinematic latent normalizing flows. However, the use of deep normalizing flows in robotic state estimation applications such as autonomous driving and UAV pose estimation remains largely unexplored. We propose the use of our deep normalizing flow framework to represent the complex and possibly multimodal beliefs over underlying robot states, mapping complex real-world distributions to highly interpretable latent representations using more advanced deep neural network architectures.

\section{Methodology}

Normalizing flows are a class of probabilistic model used for density estimation and generative modeling. Normalizing flows represent a target distribution $p(\mathbf{x})$ by transforming an easy-to-evaluate latent distribution $p(\mathbf{z})$ through the change of variable formula
\begin{equation}
    p(\mathbf{x}) = p(\mathbf{z}) |\text{det} J_f(\mathbf{z})|^{-1} \ \text{with} \ \mathbf{z} = f^{-1}(\mathbf{x})
\end{equation}
 where $f$ is a parametric transform composed of $i$ transformations $f = f_i \circ \ldots \circ f_1$. The Jacobian $J_f$ of $f$ must be invertible and differentiable; its determinant is a volume-correcting term for the probability density function of the transformed variable. The flow parameters are optimized by minimizing the KL divergence between the target distribution and the latent space under the transform $f$.

\subsection{Autoregressive Normalizing Flows}

\begin{figure}
\centering
  \centering

    \includegraphics[width=0.6\linewidth]{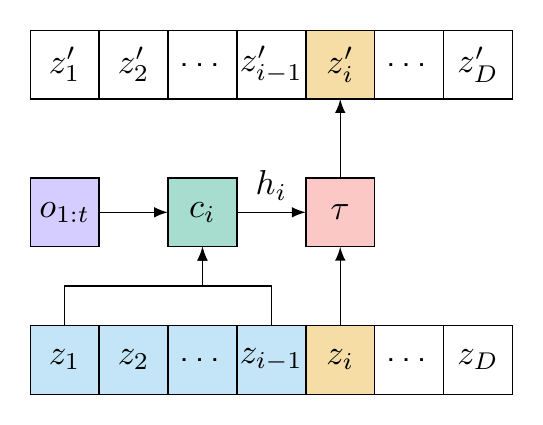}
    \caption{The $i$-th step of our conditional autoregressive normalizing flow, as adapted from \cite{papamakarios2021normalizinginference}. The conditioning operator $c_i$ takes as input latent variables $z$ and observation sequence $o_{1:t}$. The transformation operator $\tau$  uses the output $h_i$ to transform latent variables $z_i$.}
    \label{fig:af-flow}

\vspace{5mm}

  \centering
    \includegraphics[width=0.95\linewidth]{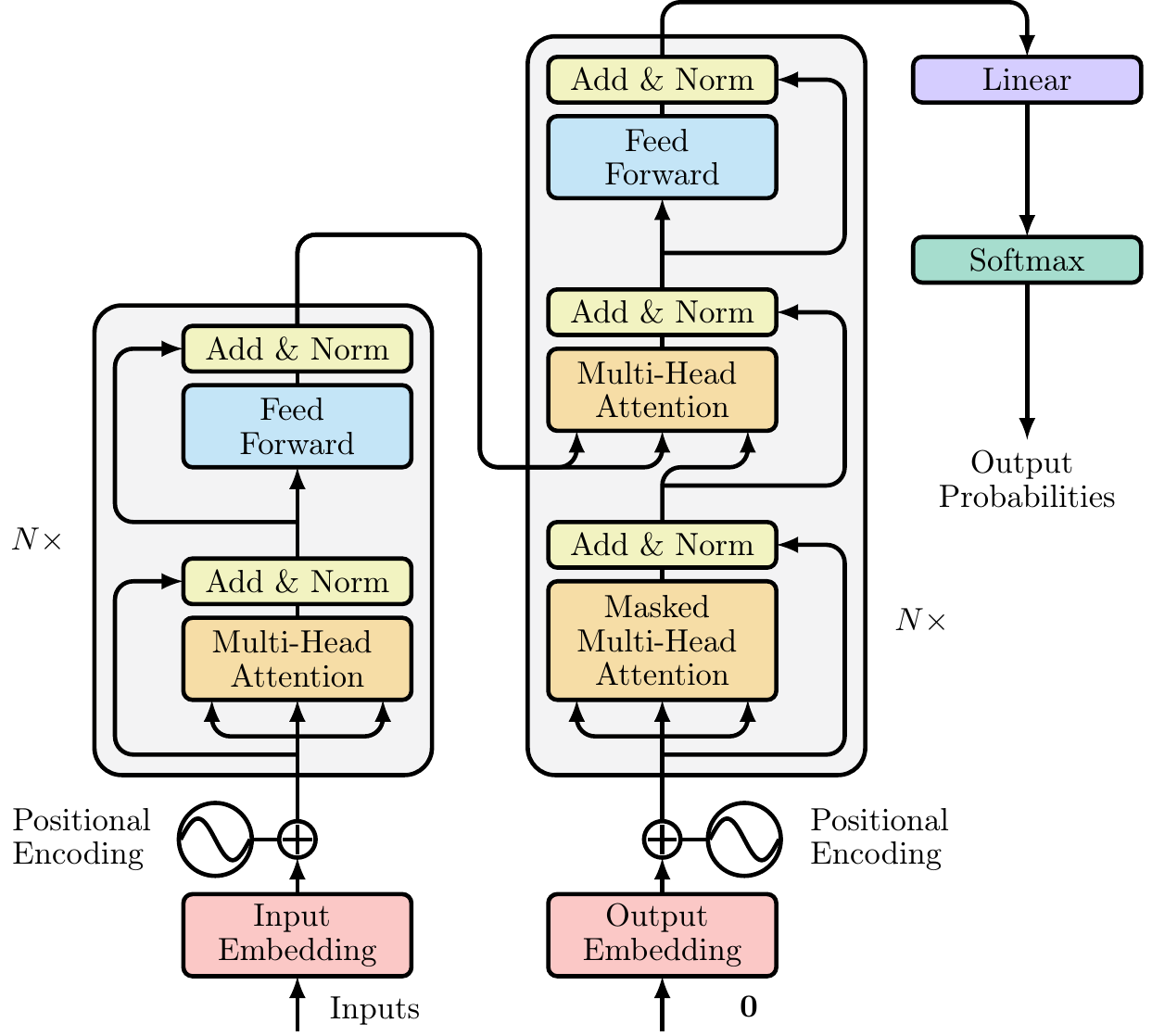}
    \caption{Transformer architecture, as adapted from \cite{vaswani_attention_2017}. We propose the use of transformers for the conditioning operator in a normalizing flow.}
    \label{fig:transformer}
\end{figure}

Autoregressive flows consist of transformations constructed in an autoregressive manner. For a single transformation, the transform of the $i$-th component of $\mathbf{z}$ to $\mathbf{z'}$ is
\begin{equation}
    z'_i = \tau(z_i, \mathbf{h}_i) \ \text{with} \ \mathbf{h}_i=c_i(\mathbf{z}_{<i})
\end{equation}
where $\tau$ is the \textit{transformation operator}, parameterized by $\mathbf{h}_i$, and $c_i$ is the \textit{conditioning operator}. This step is illustrated in \cref{fig:af-flow}. 

Normalizing flows require a tractable inverse to mimic the GAN and VAE functionality of sampling from the distribution. Thus, the transformation operator must be invertible. However, no invertibility requirement exists for the conditioning operator, which merely parameterizes the transformation operator. Thus, we are encouraged to explore various deep neural network architectures for the conditioning operator.

\subsection{Flow Architecture}
Our framework consists of stacked layers of \textit{permutation}, \textit{linear}, and \textit{masked affine autoregressive} \cite{papamakarios2017masked} flows. Permuting input variables between flow layers has been shown to help the model learn the target transformation \cite{papamakarios2021normalizinginference}. A linear flow is an invertible linear transformation of the form $\mathbf{z} = \mathbf{Wx}$, where $\mathbf{W}$ is an invertible matrix that parameterizes the transformation. Finally, a masked affine autoregressive layer in the style of \cref{fig:af-flow} passes outputs to subsequent layers. A deep conditioning operator conditions the flow on the provided \textit{context}, while a context encoder encodes the conditional information to the latent distribution parameters. The context represents the \textit{measurement} in the state estimation literature.

\subsection{Conditioner Scheme} \label{chap:conditioner}
The $i$-th conditioning operator takes $\mathbf{z}_{1:i-1}$ and $\mathbf{o}_{1:t}$ as an input and outputs the parameters $\mathbf{h}_i$ that parameterize the transformation operator $\tau$. In a naive approach, we could train $i$ different conditioners; however, this is inefficient as all conditioners have a similar task. Using a single conditioning operator for all $i$ flow transformations requires deep architectures that take sequential inputs such as recurrent neural networks (RNN) \cite{rumelhart1985learning}, gated recurrent units (GRU) \cite{cho2014properties}, long short-term memory (LSTM) \cite{hochreiter_long_1997}, or transformers \cite{vaswani_attention_2017}. Sequences of observations $\mathbf{o}_{1:t}$, possibly of varying length, necessitate a recurrent architecture.

\section{Dataset Generation}

\begin{figure}
\centering
\includegraphics[width=\linewidth]{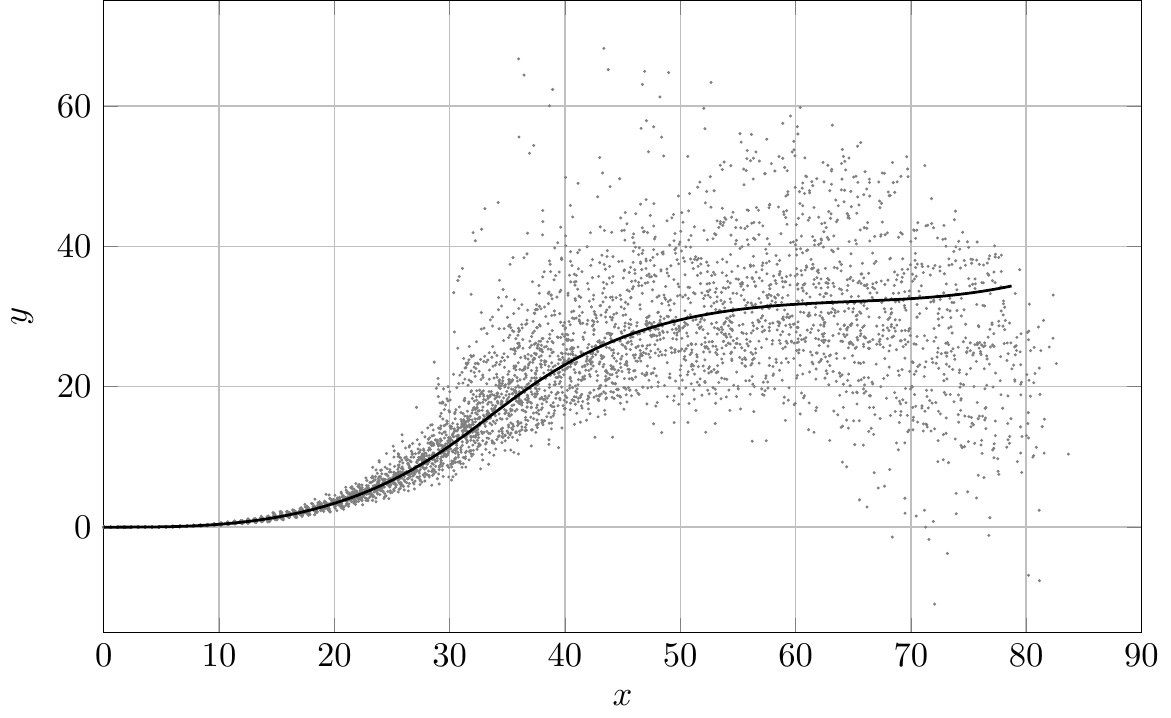}
\caption{Unimodal driving dataset}
\label{fig:unimodal_data}
\end{figure}

\begin{figure}
\centering
\includegraphics[width=\linewidth]{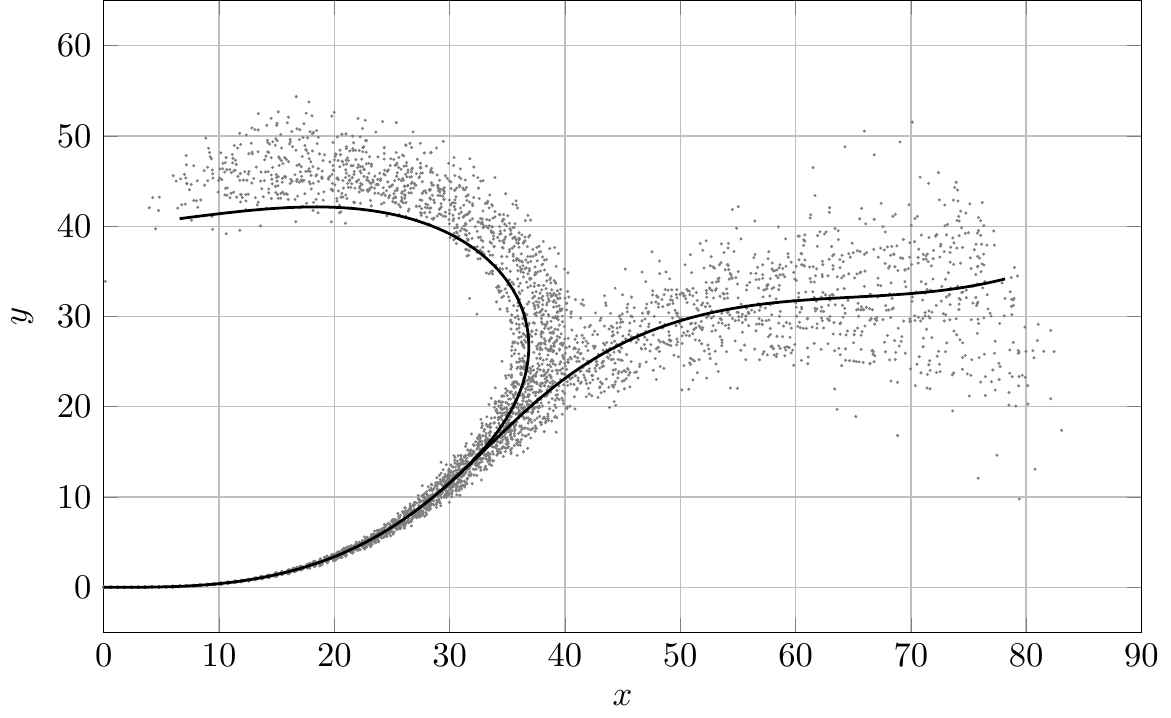}
\caption{Bimodal driving dataset}
\label{fig:bimodal_data}
\end{figure}

We construct two robotic driving datasets in simulation to validate the proposed architecture. Consider a nonholonomic robot with position $(p_x,~p_y)$ and heading angle $\theta$, and control over its velocity $v$ and angular acceleration $\phi$. The robot's equations of motion are given by 
\begin{align*}
  p_x^{t+1} &= p_x^{t} + \Delta t \cdot v^t \cdot \cos(\theta^t) \\
  p_y^{t+1} &= p_y^{t} + \Delta t \cdot v^t \cdot \sin(\theta^t) \\
  \theta^{t+1} &= \theta^{t} + \Delta t v^t \phi^t 
\end{align*}
The angular acceleration is given by
\begin{align*}
  \phi^{t+1} &= \phi^{t} + \Delta t \cdot \psi \cdot c_1 \cdot \cos(c_2 \cdot t)
\end{align*}
where $c_1$ and $c_2$ are constants and $\psi$ is a \textit{switching parameter} used to induce multimodalities into the dataset. Time-stamped coordinates from individual rollouts $(p_x, p_y, t)$ with noise-corrupted inputs $v$ and $\phi$ are shown in \cref{fig:unimodal_data,fig:bimodal_data}. 
The black lines indicate the nominal noise-free trajectory. A unimodal dataset, shown in \cref{fig:unimodal_data}, is generated by fixing $\psi = 1$ for every rollout. A bimodal dataset, possibly representing trajectories through a traffic circle, is shown in \cref{fig:bimodal_data}. It is generated by randomly selecting $\psi \sim \mathcal{U}(\left[-1, 1\right])$ at a fixed time index and thereafter holding $\psi$ constant for the remainder of the rollout.
Each dataset contains over $1.5$ million data points.

\section{Experimental Results}

\begin{figure*}[t]
    \begin{subfigure}[t]{0.48\textwidth}
    \centering
        \includegraphics[width=\linewidth]{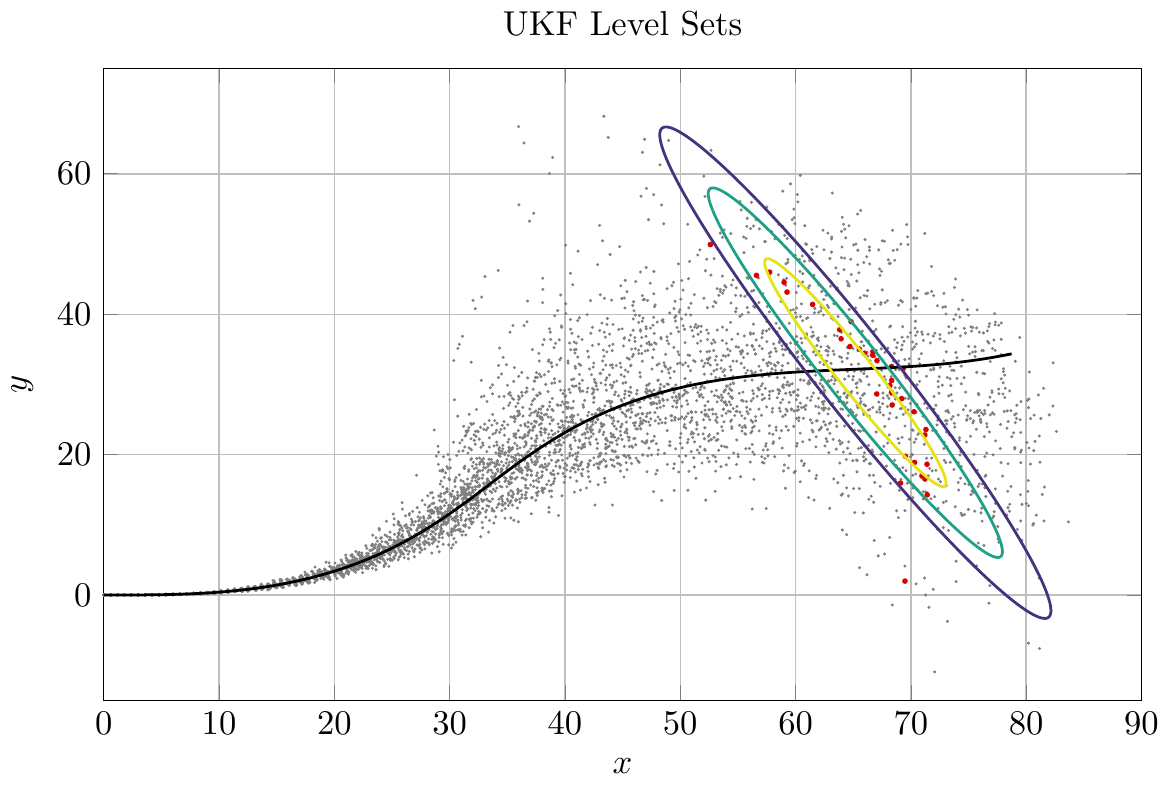}
        \label{fig:ukf_level_sets}
    \end{subfigure}
    \hfill
    \begin{subfigure}[t]{0.48\textwidth}
    \centering
        \includegraphics[width=\linewidth]{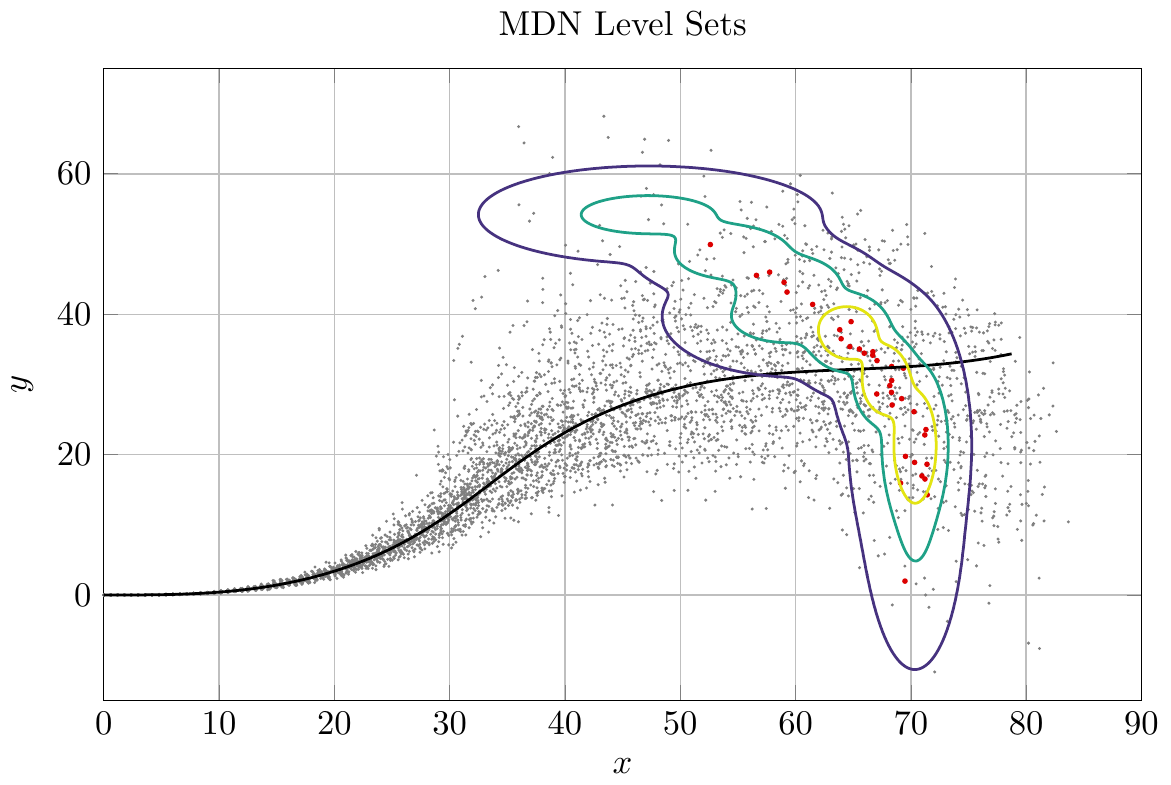}
        \label{fig:mdn_level_sets}
    \end{subfigure}
    \hfill
        \begin{subfigure}[t]{0.48\textwidth}
    \centering
        \includegraphics[width=\linewidth]{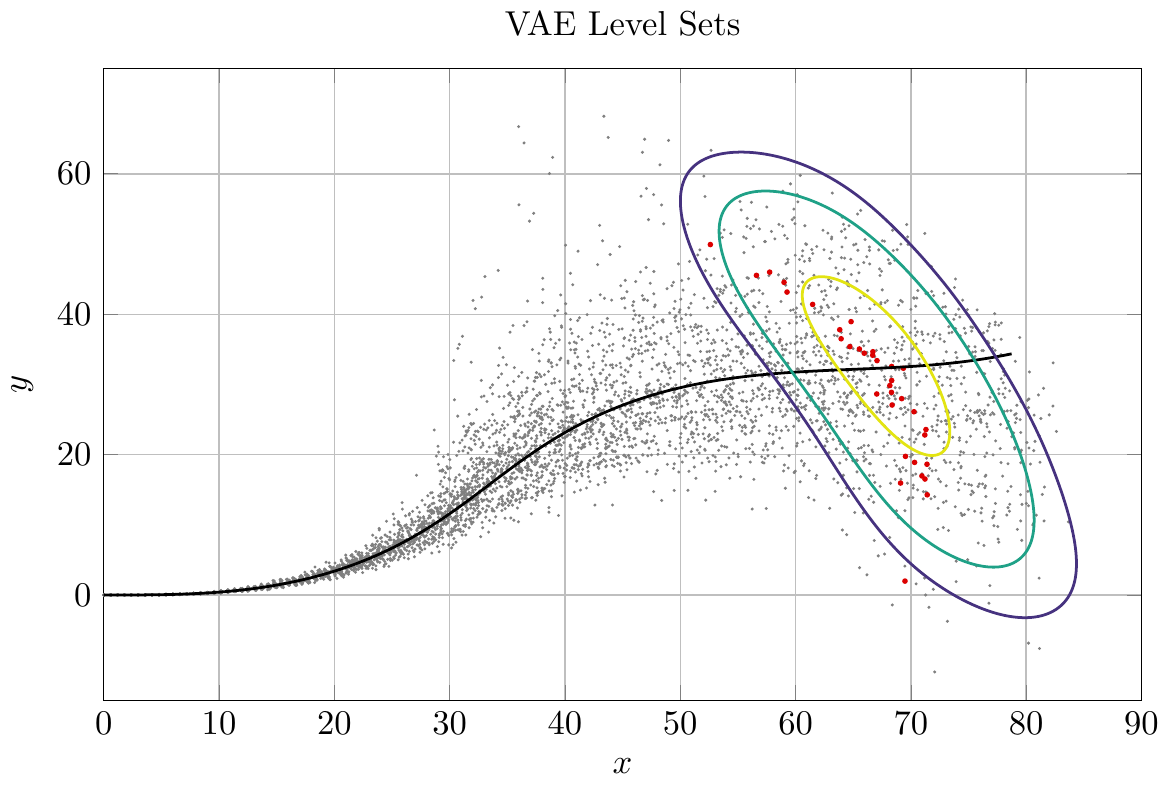}
        \label{fig:vae_level_sets}
    \end{subfigure}
    \hfill
    \begin{subfigure}[t]{0.48\textwidth}
    \centering
        \includegraphics[width=\linewidth]{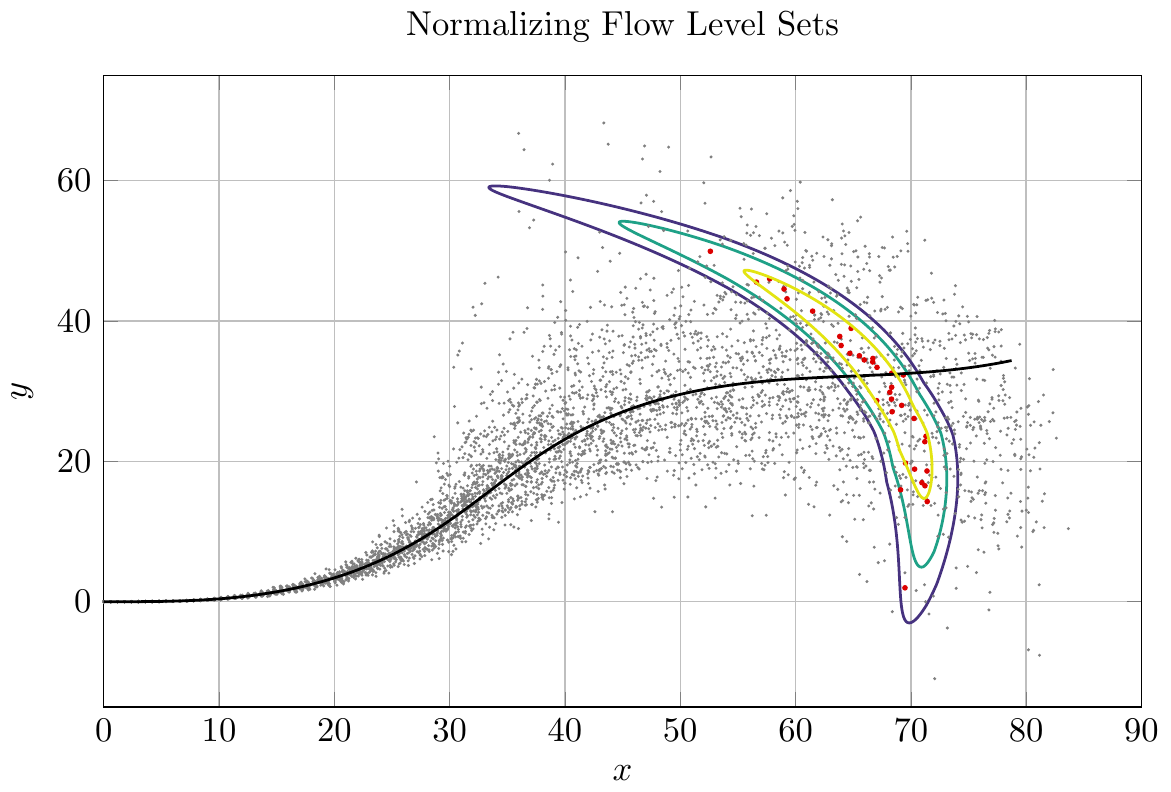}
        \label{fig:nflow_level_sets}
    \end{subfigure}
    \hfill
\caption{Confidence regions of baseline models and normalizing flows conditioned for $t=\SI{13}{\second}$. The yellow ($1\sigma$), turquoise ($2\sigma$), and the purple ($3\sigma$) contours represent confidence regions of the model. The red dots are datapoints at $t=\SI{13}{\second}$.}
\label{fig:level_sets}
\end{figure*}

We demonstrate the representational capabilities of a deep normalizing flow-based approach on two state-estimation tasks using our autonomous driving datasets. The source code is available online.\footnote{\url{ https://github.com/sisl/DeepNFStateEstimation.git}} 

\subsection{Temporal Conditioning Task}
We first consider the unimodal dataset and condition on the time stamp of each data point, estimating the conditional distribution $p(p_x, p_y \mid t)$. An MLP context encoder encodes the conditional information to the parameters of a latent conditional diagonal normal distribution, while the conditioning operator is simply the identity function. We benchmark our proposed approach against three baselines:

\begin{itemize}
    \item \textbf{Unscented Kalman Filter}: A UKF is a classical recursive Bayesian filter that relies on a deterministic sampling strategy to approximate the effect of a distribution undergoing a nonlinear transformation \cite{julier1997new}. 
    \item \textbf{Mixture Density Network}: An MDN represents a conditional Gaussian mixture model where the components are parameterized by a neural network. We use an MDN with 5 mixture components parameterized by a two-layer network with 8 neurons per layer and tanh activations.
    \item \textbf{VAE}: We use a conditional VAE architecture with identical encoder and decoder architectures, each consisting of 8 hidden layers with 64 neurons each and tanh as the activation function.
\end{itemize}

The results of the baseline methods and our current implementation of normalizing flows are depicted in \cref{fig:level_sets} where we show the $1\sigma$, $2\sigma$, and $3\sigma$ confidence regions. Additionally, we estimate the KL divergence between the dataset and $\num{1000}$ samples drawn from the fitted models using a two-sample KL divergence estimator \cite{perez2008kullback}. Given $n$ i.i.d. samples $\in \mathbb{R}^d$ from the flow density $\hat{p}(\mathbf{z})$ and $m$ i.i.d. samples $\in \mathbb{R}^d$ from the target density $p(\mathbf{z})$, an approximation of the KL divergence is given by
\begin{equation}
\label{eq:approx-dkl}
    \hat{D}_\text{KL}\infdivx{\hat{p}(\mathbf{x})}{p(\mathbf{z})} = \dfrac{d}{n} \sum_{i=1}^n \log \dfrac{r_k(\mathbf{z}_i)}{s_k(\mathbf{z}_i)} + \log \dfrac{m}{n-1}
\end{equation}
where $r_k(\mathbf{z}_i)$ and $s_k(\mathbf{z}_i)$ are the Euclidean distance to the $k^\text{th}$ nearest-neighbor of $\mathbf{z}_i$ in the samples from $\hat{p}(\mathbf{z})$ and $p(\mathbf{z})$, respectively. Approximate KL divergence measures are presented in Table \ref{tab:kl_est}. Our normalizing flow implementation outperformed the baselines, as supported by both visual evidence and quantitative analysis.

\begin{table}[h]
\centering
\caption{Estimated KL divergence between models and dataset. Lower KL divergence indicates better performance.\label{tab:kl_est}}
\begin{tabular}{@{}lr@{}}
\toprule
\bf{Method} & $\hat{D}_\text{KL}\infdivx{\hat{p}(\mathbf{x})}{p(\mathbf{z})}$ \\ \midrule
Unscented Kalman Filter    &    0.7578                                                       \\
Mixture Density Network    &    0.2612                                                     \\
Variational Autoencoder    &    0.7978                                                      \\ 
\textbf{Normalizing Flows}    &  \textbf{0.0684}                                                       \\ \bottomrule
\end{tabular}
\vspace*{0.15cm}
\end{table}

\subsection{Observation History Conditioning Task}
For the second task we condition on sequences of noisy position observations and estimate the next state of the vehicle; that is, we estimate the conditional distribution $p(\mathbf{x}_{t} \mid \mathbf{o}_{1:t-1})$. Due to the time-series nature of the data, we consider recurrent conditioning operators. The following conditioning operator architectures are tested:

\begin{itemize}
    \item \textbf{Recurrent Neural Network}: A single-layer RNN with three input features and four output features serves as a baseline.
    \item \textbf{Gated Recurrent Unit}: A single-layer GRU with three input features, four hidden features, and four output features is employed to improve upon the RNN baseline by solving the vanishing gradient problem.
    \item \textbf{Long Short-Term Memory}: A single-layer LSTM with three input features, four hidden features, and four output features is employed to improve upon the RNN baseline by solving the vanishing gradient problem while providing more expressivity than the GRU.
    \item \textbf{Transformer}: We use a transformer network and follow the same architecture as \cite{wu2020deep}. Since we use the transformer to learn an embedding, no ground truth for the target sequence is available. This differs from work by \citet{rasul2020multivariate}, who use future observations as input to the decoder. This approach enables us to freely choose the dimensionality of the embedding space. As learning an embedding is a many-to-one task, only one prediction step in ``inference'' mode is necessary. We initialize the target sequence required by the transformer with zeros, which is common practice for other recurrent methods. 
\end{itemize}

\begin{figure*}
\input{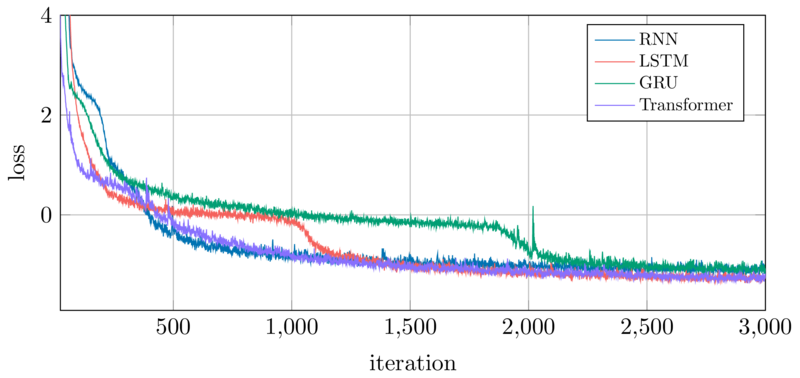}
\captionof{figure}{Training loss for normalizing flows based on different embedding networks for the first $3,000$ training iterations.}
\label{fig:loss_curves}
\end{figure*}

We train the normalizing flow with all four conditioning operators for $\num{10000}$ iterations with a batch size of $\num{2048}$. The Adam optimizer with default parameters is used for all flow configurations. The training loss is depicted in \cref{fig:loss_curves}.

\begin{table}[h]
\centering
\caption{Mean log likelihood of undisturbed data points under the learned distribution, represented by the normalizing flow. Higher log likelihood indicates better performance.\label{tab:log_prob}}
\begin{tabular}{@{}lr@{}}
\toprule
\bf{Conditioning Operator} & \bf{Log Likelihood} \\ \midrule
RNN    &    1.4255  \\
GRU    &    1.4818  \\
LSTM    &    1.5057 \\ 
\textbf{Transformer}    &  \textbf{1.5736}  \\ 
\bottomrule
\end{tabular}
\end{table}

\begin{figure*}[h]
    \begin{subfigure}[t]{0.48\textwidth}
    \centering
        \includegraphics[width=\linewidth]{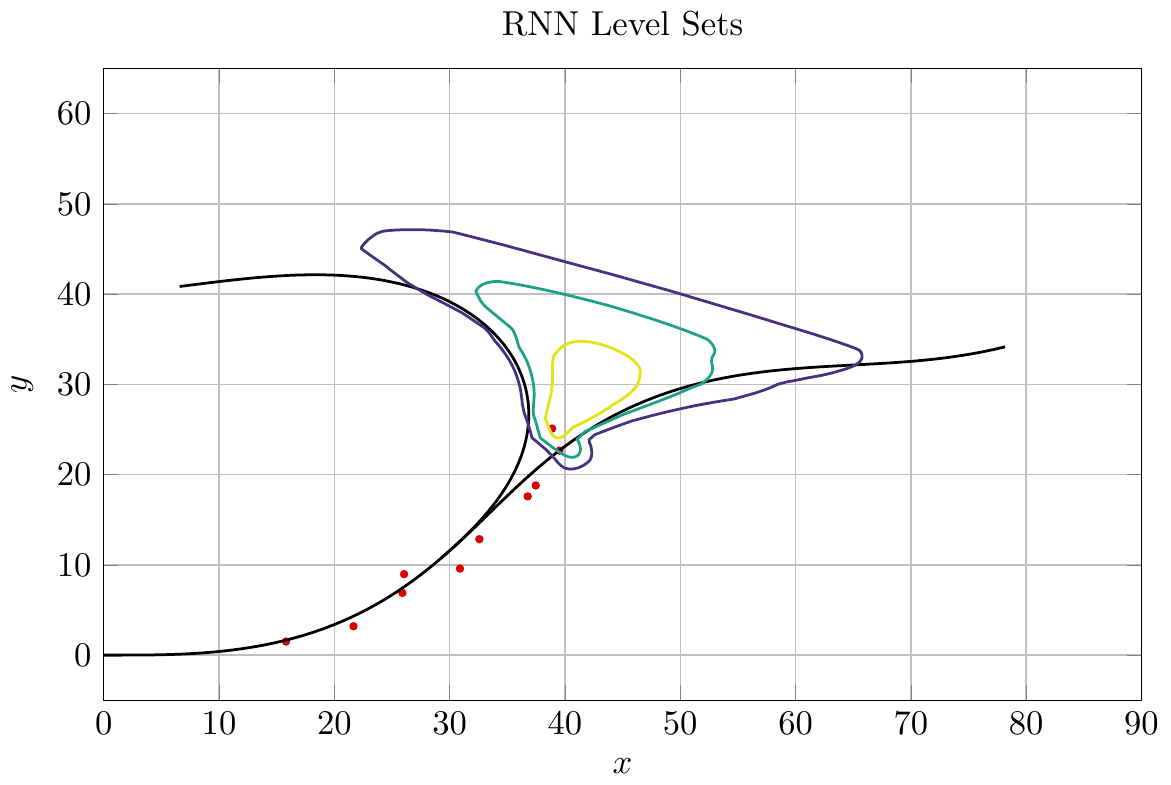}
        \label{fig:rnn_level_sets}
    \end{subfigure}
    \hfill
    \begin{subfigure}[t]{0.48\textwidth}
    \centering
        \includegraphics[width=\linewidth]{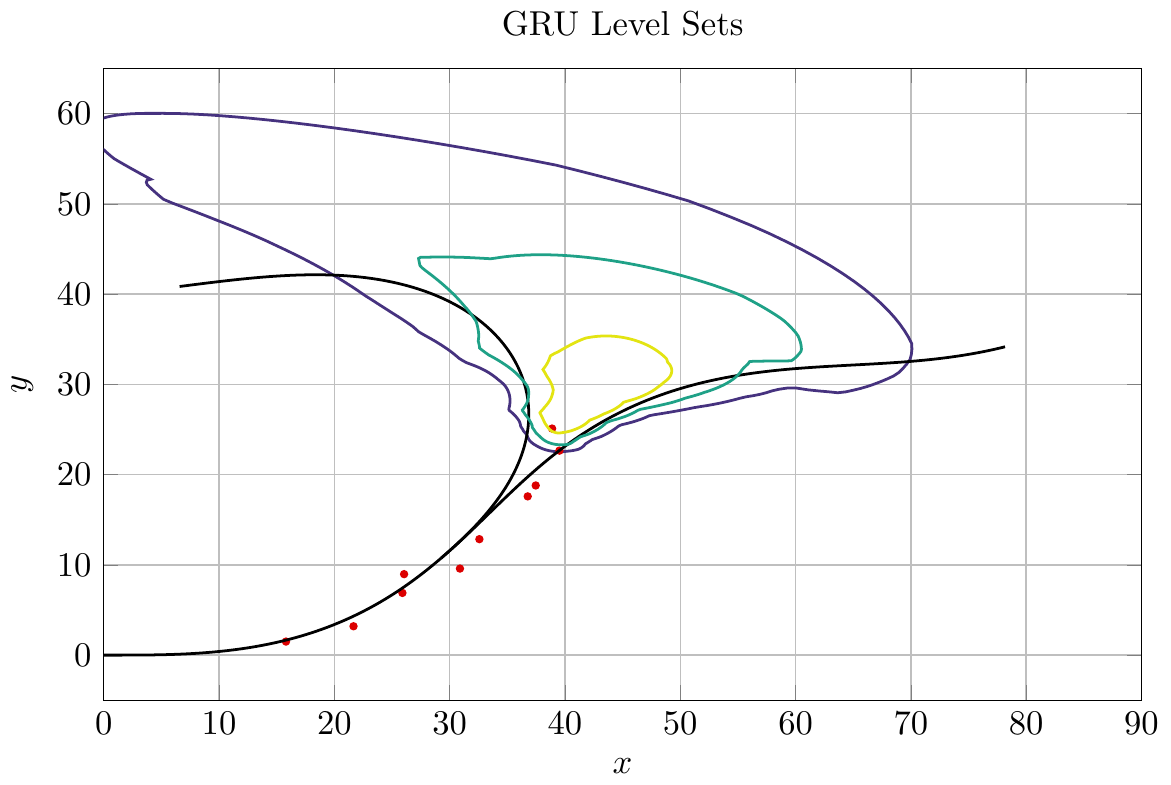}
        \label{fig:gru_level_sets}
    \end{subfigure}
    \hfill
        \begin{subfigure}[t]{0.48\textwidth}
    \centering
        \includegraphics[width=\linewidth]{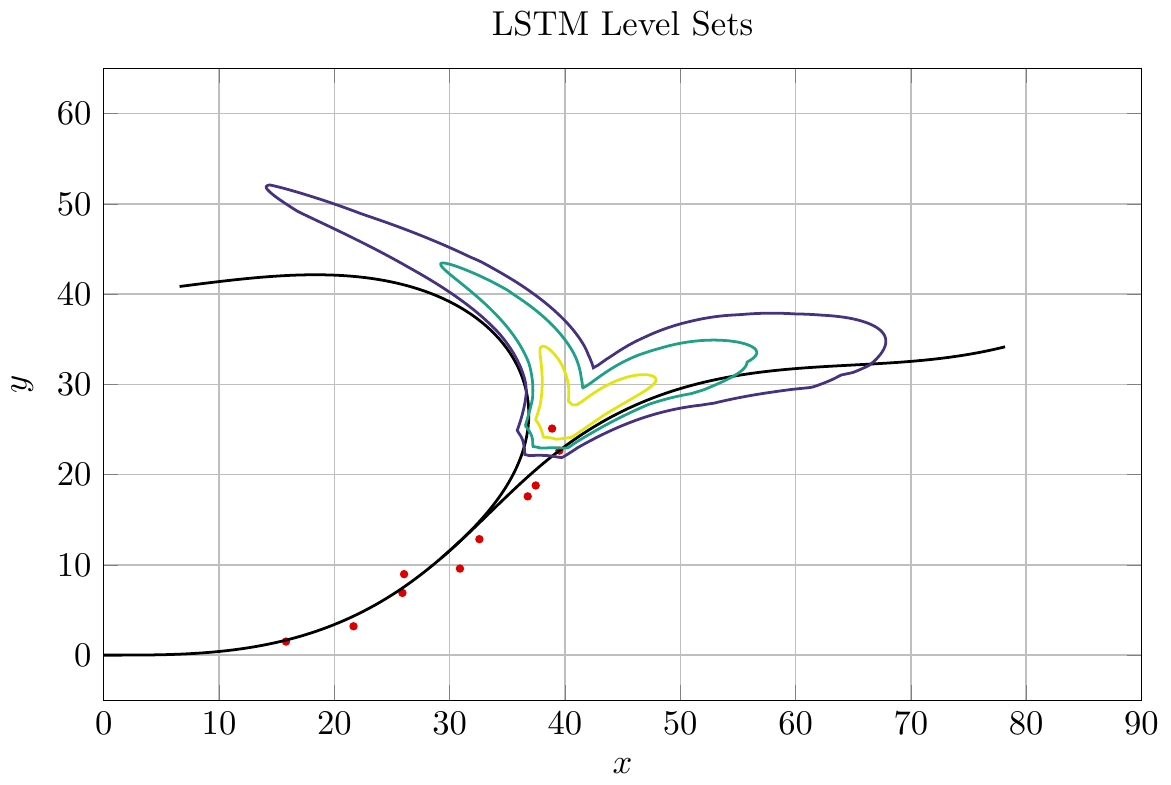}
        \label{fig:lstm_level_sets}
    \end{subfigure}
    \hfill
    \begin{subfigure}[t]{0.48\textwidth}
    \centering
        \includegraphics[width=\linewidth]{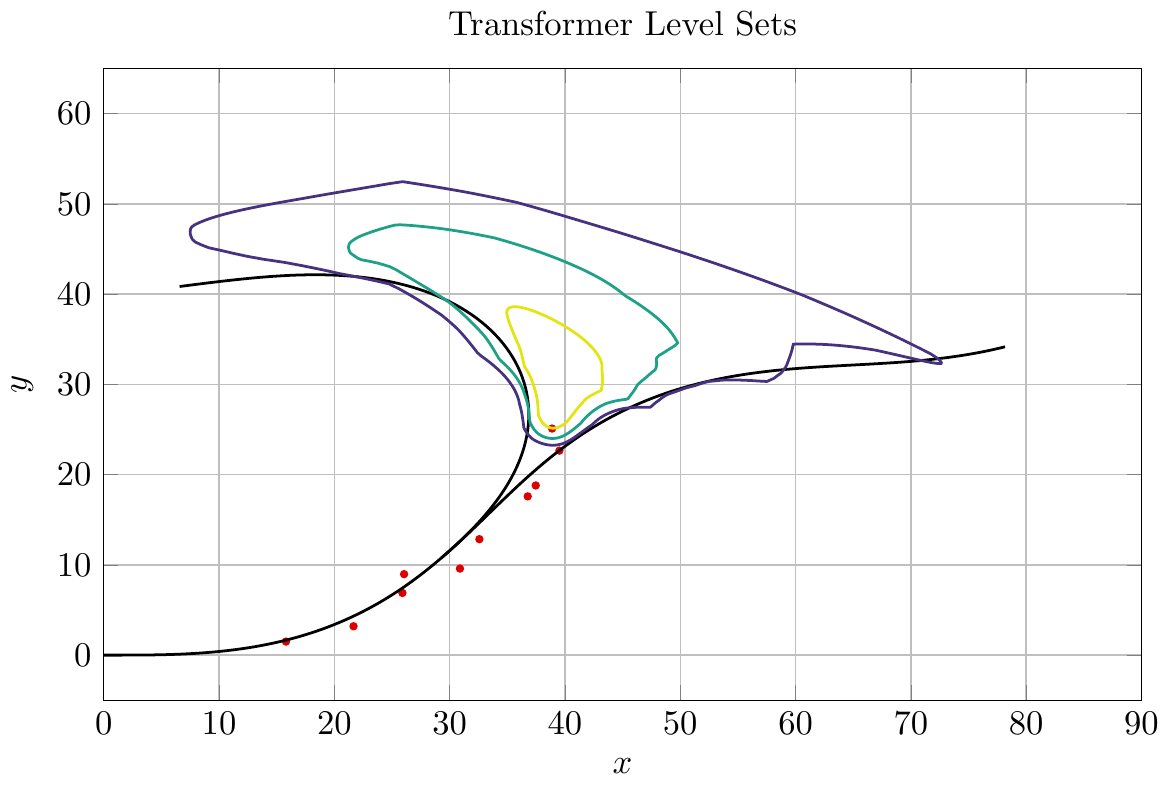}
        \label{fig:transformer_level_sets}
    \end{subfigure}
    \hfill
\caption{Confidence regions of four conditioning network architectures conditioned on a sequence of noisy observations. The yellow ($1\sigma$), turquoise ($2\sigma$), and the purple ($3\sigma$) contours represent confidence regions of the model. The red dots are noisy observations for a single trajectory. The trajectory is a left-branching trajectory. Both RNNs and GRU predict a rather right-branching trajectory, the LSTM is indecisive, and only the transformer has a slightly left-branching prediction.}
\label{fig:bimodal_level_sets}
\end{figure*}

The results of the four deep recurrent conditioning operator architectures are depicted in \cref{fig:bimodal_level_sets} where we show the $1\sigma$, $2\sigma$, and $3\sigma$ confidence regions over the predicted next state. Additionally, we compute the mean log likelihood of the ground truth positions given the associated context. The model attempts to maximize the log likelihood of the noise-free data points, and thus a higher mean log likelihood indicates that the learned distribution more closely matches the true underlying belief. The transformer architecture yields the highest mean log likelihood value, as shown in \cref{tab:log_prob}.

\section{Discussion and Future Work}

We evaluate four deep conditioning operator architectures for conditioning a normalizing flow and an MLP context encoder for encoding conditional information to the parameters of a latent base distribution. Normalizing flows outperformed both classical state-estimation techniques and deep conditional networks on a simple state-estimation task---even when an identity function conditioning operator was used---due to their expressive representational capabilities. Furthermore, we demonstrated that deep recurrent conditioning operator architectures are well-suited for time-series contexts on a complex state-estimation task with a bimodal underlying belief distribution. A transformer network was experimentally shown to provide the best quantitative results in the proposed framework. This appears to be the first time that a transformer architecture has been used as a normalizing flow conditioning operator, and represents one of the first applications of normalizing flows to state-estimation tasks.

In future work we will consider alternative flow architectures such as Unconstrained Monotonic Neural Network autoregressive flows \cite{wehenkel2019unconstrained}. Furthermore, additional conditioning operator architectures---such as diffusion models---will be considered. Finally, we will validate our deep normalizing flow framework on high-dimensional state-estimation tasks such as UAV pose estimation. Our approach may be able to use the high-dimensional environment information in the network's conditioning operator.

\section*{Acknowledgments}
This research was supported in part by the National Science Foundation Graduate Research Fellowship Program under Grant No. DGE-1656518 and Grant No. DGE-2146755.

\renewcommand*{\bibfont}{\footnotesize}
\printbibliography

@String { cvpr        = {IEEE Computer Society Conference on Computer Vision and Pattern Recognition (CVPR)} }

@String { icassp      = {International Conference on Acoustics, Speech, and Signal Processing (ICASSP)} }

@String { itsc        = {IEEE International Conference on Intelligent Transportation Systems (ITSC)} }

@String { jmlr        = {Journal of Machine Learning Research} }

@String { nips        = {Advances in Neural Information Processing Systems (NIPS)} }

@String { neurips     = {Advances in Neural Information Processing Systems (NeurIPS)} }

@article{kobyzevintroductionmethods,
  author={Kobyzev, Ivan and Prince, Simon J.D. and Brubaker, Marcus A.},
  journal={IEEE Transactions on Pattern Analysis and Machine Intelligence}, 
  title={Normalizing Flows: An Introduction and Review of Current Methods}, 
  year={2021},
  volume={43},
  number={11},
  pages={3964-3979},
  doi={10.1109/TPAMI.2020.2992934}}

@article{papamakarios2021normalizinginference,
  title={Normalizing Flows for Probabilistic Modeling and Inference},
  author={Papamakarios, George and Nalisnick, Eric T and Rezende, Danilo Jimenez and Mohamed, Shakir and Lakshminarayanan, Balaji},
  journal=jmlr,
  volume={22},
  number={57},
  pages={1--64},
  year={2021}
}

@inproceedings{vaswani_attention_2017,
	title = {Attention is {All} you {Need}},
	volume = {30},
	url = {https://proceedings.neurips.cc/paper/2017/hash/3f5ee243547dee91fbd053c1c4a845aa-Abstract.html},
	urldate = {2022-10-12},
	booktitle = neurips,
	author = {Vaswani, Ashish and Shazeer, Noam and Parmar, Niki and Uszkoreit, Jakob and Jones, Llion and Gomez, Aidan N and Kaiser, Łukasz and Polosukhin, Illia},
	year = {2017},
}

@article{hochreiter_long_1997,
	title = {Long {Short}-{Term} {Memory}},
	volume = {9},
	issn = {0899-7667},
	doi = {10.1162/neco.1997.9.8.1735},
	number = {8},
	journal = {Neural Computation},
	author = {Hochreiter, Sepp and Schmidhuber, Jürgen},
	month = nov,
	year = {1997},
	pages = {1735--1780},
}

@techreport{rumelhart1985learning,
  title={Learning internal representations by error propagation},
  author={Rumelhart, David E and Hinton, Geoffrey E and Williams, Ronald J},
  year={1985},
  institution={California Univ San Diego La Jolla Inst for Cognitive Science},
  number={ADA164453}
}

@article{cho2014properties,
  title={On the properties of neural machine translation: Encoder-decoder approaches},
  author={Cho, Kyunghyun and Van Merri{\"e}nboer, Bart and Bahdanau, Dzmitry and Bengio, Yoshua},
  journal={arXiv preprint arXiv:1409.1259},
  year={2014}
}

@book{barfoot2017state,
  title={State estimation for robotics},
  author={Barfoot, Timothy D},
  year={2017},
  publisher={Cambridge University Press}
}

@article{bishop1994mixture,
  title={Mixture density networks},
  author={Bishop, Christopher M},
  year={1994},
  publisher={Aston University}
}

@article{kalman1960new,
  title={A new approach to linear filtering and prediction problems},
  author={Kalman, Rudolph Emil},
  year={1960}
}

@inproceedings{julier1997new,
  title={New extension of the {K}alman filter to nonlinear systems},
  author={Julier, Simon J and Uhlmann, Jeffrey K},
  booktitle={Signal Processing, Sensor Fusion, and Target Recognition VI},
  volume={3068},
  pages={182--193},
  year={1997},
  organization={SPIE}
}

@article{goodfellow2020generative,
  title={Generative adversarial networks},
  author={Goodfellow, Ian and Pouget-Abadie, Jean and Mirza, Mehdi and Xu, Bing and Warde-Farley, David and Ozair, Sherjil and Courville, Aaron and Bengio, Yoshua},
  journal={Communications of the ACM},
  volume={63},
  number={11},
  pages={139--144},
  year={2020},
  publisher={ACM New York, NY, USA}
}

@article{kingma2013auto,
  title={Auto-encoding variational Bayes},
  author={Kingma, Diederik P and Welling, Max},
  journal={arXiv preprint arXiv:1312.6114},
  year={2013}
}

@article{deng2020modeling,
  title={Modeling continuous stochastic processes with dynamic normalizing flows},
  author={Deng, Ruizhi and Chang, Bo and Brubaker, Marcus A and Mori, Greg and Lehrmann, Andreas},
  journal=neurips,
  volume={33},
  pages={7805--7815},
  year={2020}
}

@inproceedings{ma2020normalizing,
  title={Normalizing flow policies for multi-agent systems},
  author={Ma, Xiaobai and Gupta, Jayesh K and Kochenderfer, Mykel J},
  booktitle={International Conference on Decision and Game Theory for Security},
  pages={277--296},
  year={2020},
  organization={Springer}
}

@inproceedings{zhang2022accelerated,
  title={Accelerated Testing for Highly Automated Vehicles: A Combined Method Based on Importance Sampling and Normalizing Flows},
  author={Zhang, He and Sun, Jian and Tian, Ye},
  booktitle=itsc,
  pages={574--579},
  year={2022},
  organization={IEEE}
}

@inproceedings{zanfir2020weakly,
  title={Weakly supervised 3d human pose and shape reconstruction with normalizing flows},
  author={Zanfir, Andrei and Bazavan, Eduard Gabriel and Xu, Hongyi and Freeman, William T and Sukthankar, Rahul and Sminchisescu, Cristian},
  booktitle={European Conference on Computer Vision},
  pages={465--481},
  year={2020},
  organization={Springer}
}

@inproceedings{li2019generating,
  title={Generating multiple hypotheses for 3d human pose estimation with mixture density network},
  author={Li, Chen and Lee, Gim Hee},
  booktitle=cvpr,
  pages={9887--9895},
  year={2019}
}

@INPROCEEDINGS{zen2014speech,
  author={Zen, Heiga and Senior, Andrew},
  booktitle=icassp, 
  title={Deep mixture density networks for acoustic modeling in statistical parametric speech synthesis}, 
  year={2014},
  volume={},
  number={},
  pages={3844-3848},
  }

@InProceedings{Makansi_2019_CVPR,
    author = {Makansi, Osama and Ilg, Eddy and Cicek, Ozgun and Brox, Thomas},
    title = {Overcoming Limitations of Mixture Density Networks: A Sampling and Fitting Framework for Multimodal Future Prediction},
    booktitle = cvpr,
    month = jun,
    year = {2019}
}

@article{rocchetto2018learning,
  title={Learning hard quantum distributions with variational autoencoders},
  author={Rocchetto, Andrea and Grant, Edward and Strelchuk, Sergii and Carleo, Giuseppe and Severini, Simone},
  journal={NPJ Quantum Information},
  volume={4},
  number={1},
  pages={1--7},
  year={2018},
  publisher={Nature Publishing Group}
}

@inproceedings{bethge2022eeg2vec,
  title={EEG2Vec: Learning Affective EEG Representations via Variational Autoencoders}, 
  author={Bethge, David and Hallgarten, Philipp and Grosse-Puppendahl, Tobias and Kari, Mohamed and Chuang, Lewis L. and Özdenizci, Ozan and Schmidt, Albrecht},
  booktitle={IEEE International Conference on Systems, Man, and Cybernetics (SMC)}, 
  year={2022},
  pages={3150-3157},
  doi={10.1109/SMC53654.2022.9945517}
}

@inproceedings{perez2008kullback,
  title={Kullback-Leibler divergence estimation of continuous distributions},
  author={P{\'e}rez-Cruz, Fernando},
  booktitle={IEEE International Symposium on Information Theory},
  pages={1666--1670},
  year={2008},
  organization={IEEE}
}

@article{wu2020deep,
  title={Deep transformer models for time series forecasting: The influenza prevalence case},
  author={Wu, Neo and Green, Bradley and Ben, Xue and O'Banion, Shawn},
  journal={arXiv preprint arXiv:2001.08317},
  year={2020}
}

@article{wehenkel2019unconstrained,
  title={Unconstrained monotonic neural networks},
  author={Wehenkel, Antoine and Louppe, Gilles},
  journal=neurips,
  volume={32},
  year={2019}
}

@article{rasul2020multivariate,
  title={Multivariate probabilistic time series forecasting via conditioned normalizing flows},
  author={Rasul, Kashif and Sheikh, Abdul-Saboor and Schuster, Ingmar and Bergmann, Urs and Vollgraf, Roland},
  journal={arXiv preprint arXiv:2002.06103},
  year={2020}
}

@article{papamakarios2017masked,
  title={Masked autoregressive flow for density estimation},
  author={Papamakarios, George and Pavlakou, Theo and Murray, Iain},
  journal=nips,
  volume={30},
  year={2017}
}

\end{document}